\documentclass[11pt]{article}

\usepackage[preprint]{acl}

\usepackage{times}
\usepackage{latexsym}

\usepackage[T1]{fontenc}

\usepackage[utf8]{inputenc}

\usepackage{microtype}

\usepackage{inconsolata}

\usepackage{graphicx}
\usepackage{booktabs}
\usepackage{amsmath}
\usepackage{amssymb}
\usepackage{algorithm}
\usepackage{algpseudocode}
\algrenewcommand\algorithmicrequire{\textbf{Input:}}
\algrenewcommand\algorithmicensure{\textbf{Output:}}
\title{WA-SpecDec: World-Aware Speculative Decoding for Vision-Language-Action Models}

\author{
Zikang Wen \and
Yuning Zhang \and
Dong Yuan \\
The University of Sydney \\
Sydney, Australia \\
\texttt{\{zikang.wen,yuning.zhang1,dongyuan\}@sydney.edu.au}
}

\begin{document}
\maketitle
\begin{abstract}
Vision-language-action (VLA) policies generate robot controls autoregressively, making closed-loop latency dominated by repeated target-model forward passes. Speculative decoding reduces this cost by verifying blocks of draft action tokens in parallel, and recent VLA methods further relax token-level acceptance because small differences in action-token space often map to similar continuous controls. However, this relaxation remains scene-agnostic. A fixed token-distance tolerance treats the same action-token deviation as equally safe across states, although deviations that are harmless in free space can cause collisions or grasp failures near contact. We propose WA-SpecDec, a world-aware speculative decoding framework that injects world-model-derived physical scene awareness during the VLA prefill stage, producing shared world-aware prefill states for draft proposal and target verification without changing the relaxed acceptance rule. Across three state-of-the-art relaxed acceptance schemes, WA-SpecDec preserves higher task success under looser relaxation and enables longer accepted prefixes. At comparable-success operating points, WA-SpecDec achieves a $1.5\times$ matched-success speedup over VLA speculative decoding alone and reduces near-contact failure (NCF) by $18.6\%$ on average relative to the corresponding speculative baselines.
\end{abstract}

\section{Introduction}
\begin{figure}[t]
  \centering
  \includegraphics[width=0.8\columnwidth]{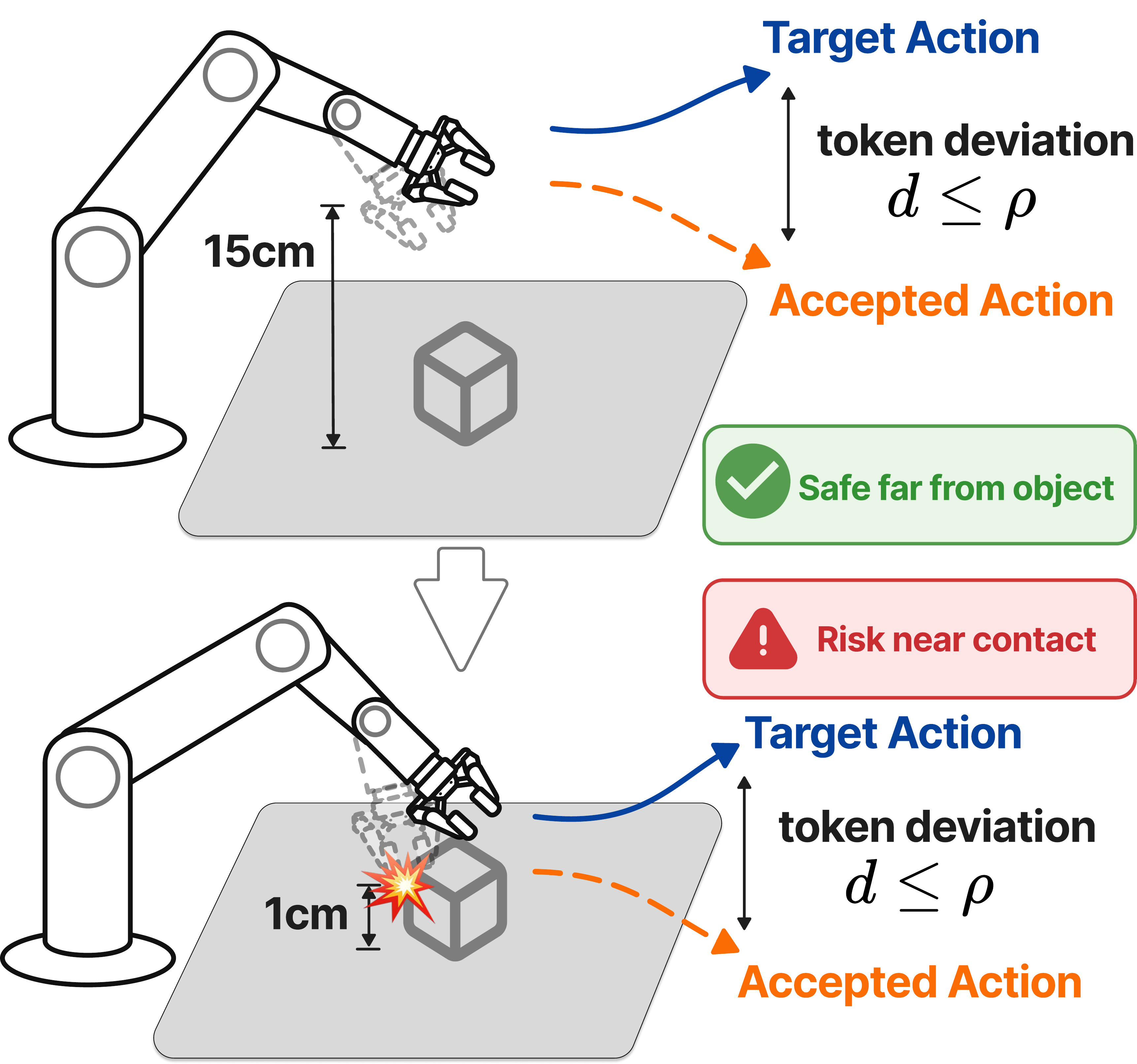}
  \caption{Motivation for world-aware speculative decoding. The same draft--target deviation can be safe in free space but risky near contact.}
  \label{fig:problem}
\end{figure}
Vision-language-action (VLA) models extend language-conditioned robot control and embodied multimodal modeling into executable robot policies \citep{pmlr-v205-ichter23a,pmlr-v202-driess23a}. Modern robot transformers map language-conditioned visual observations to action tokens or continuous commands that are decoded into robot control signals \citep{brohan2023rt,zitkovich2023rt}. Cross-embodiment datasets and open-source generalist policies further make this action-generation interface a practical foundation for embodied control \citep{10611477,pmlr-v270-kim25c}. However, token-generating VLA policies still rely on repeated target-model forward passes during autoregressive action decoding, which makes closed-loop latency a key bottleneck for fast observe--decide--act cycles.

Speculative decoding reduces this cost by using a lightweight draft model to propose multiple future tokens that are verified in parallel by the target model \citep{leviathan2023fast}. Subsequent speculative sampling and feature-aware drafting methods further improve this acceleration paradigm for autoregressive decoding \citep{chen2023accelerating,li2024eagle}. In VLA decoding, exact token agreement is often too restrictive because small differences in action-token space can map to similar continuous controls. Recent VLA speculative decoding methods therefore adopt \emph{relaxed acceptance}, accepting a draft token when it lies within a tolerance $\rho$ of the target token in action-token space \citep{wang2025spec}. Kinematic and hybrid variants further adapt this idea to embodied action spaces \citep{zheng2026kerv,zheng2026heisd}. These methods improve accepted length and wall-clock speed, but they also shift the key reliability question from exact token matching to whether an accepted token deviation is physically safe in the current scene.

Whether an accepted action-token deviation is safe is inherently scene-dependent. The same deviation can be benign during free-space motion, but near contact it may change whether the robot grasps, collides, misses the target, or enters the wrong contact phase. This creates a representational bottleneck for relaxed speculative decoding. Before action decoding, standard VLA policies construct a multimodal prefill input from language tokens and projected visual patch embeddings, which is then encoded into hidden states for autoregressive action generation. Although image-text and self-supervised pretraining provide strong semantic and spatial representations \citep{Zhai_2023_ICCV,oquab2024dinov}, these visual features are not explicitly trained to preserve predictive physical state such as object pose evolution, gripper--object distance, or contact phase \citep{nair2023r3m,radosavovic2023real}. As a result, relaxed verification applies a global token-distance test while the physical cost of an accepted deviation varies across scenes. Figure~\ref{fig:problem} illustrates this mismatch. A tolerated deviation may be safe when the gripper is far from the object, but unsafe near contact. A scene-agnostic relaxed decoder must therefore keep $\rho$ conservative to preserve task success, which limits accepted length and speedup. Looser relaxation improves acceptance, but can admit physically inconsistent actions.

To address this mismatch, we propose \emph{World-Aware Speculative Decoding for Vision-Language-Action Models} (WA-SpecDec), the first framework to introduce world-model-derived physical scene awareness into VLA speculative decoding \citep{wu2023daydreamer}. WA-SpecDec injects this awareness before prefill computation, producing world-aware prefill hidden states shared by draft proposal and target verification, rather than modifying the relaxed acceptance rule. Specifically, WA-SpecDec enriches visual patch embeddings with a World-Aware Bias module, which maps current-observation world tokens into spatially aligned biases and is trained with action prediction together with an auxiliary next-frame latent objective. At inference, it uses only the current observation and performs no future-frame prediction. By conditioning both proposal and verification on the same physically informed prefill hidden states, WA-SpecDec improves relaxed speculative decoding while preserving the original VLA action-token interface. Our contributions are threefold:
\begin{itemize}
    \item We identify a scene-dependent reliability bottleneck in relaxed VLA speculative decoding, where a fixed token-distance tolerance can accept deviations whose physical cost varies between free-space and near-contact states.
    \item We propose WA-SpecDec, the first framework that incorporates world-model-derived physical scene awareness into VLA speculative decoding by injecting a spatially aligned World-Aware Bias before prefill computation, producing shared world-aware prefill hidden states while leaving the relaxed acceptance rule unchanged.
    \item We introduce near-contact failure (NCF), a new metric to measure contact-sensitive degradation and evaluate WA-SpecDec on four LIBERO task suites across $\rho$-based, kinematic, and hybrid relaxed acceptance schemes. WA-SpecDec enables longer accepted prefixes at comparable task success, achieves a $1.5\times$ matched-success speedup, and reduces NCF by $18.6\%$ on average relative to the corresponding speculative baselines.
\end{itemize}

\section{Background and Related Work}
\label{sec:background}

\subsection{Acceleration for VLA Models}
Language-conditioned robot control first connected pretrained language models to robotic affordances \citep{pmlr-v205-ichter23a} and embodied multimodal reasoning \citep{pmlr-v202-driess23a}. Robot transformers then showed that large-scale sequence policies can map language-conditioned visual observations to executable robot actions across diverse manipulation tasks. More recent VLA and generalist robot policies extend this direction with cross-embodiment data \citep{10611477}, open-source action-token policies \citep{octo_2023, PertschK-RSS-25}, and action-chunk or flow-based control heads \citep{BlackK-RSS-25}. These systems improve broad task transfer and generalization, but token-generating VLA policies still inherit the latency profile of autoregressive decoding \citep{zitkovich2023rt,pmlr-v270-kim25c}.

At each control step $t$, a VLA policy receives a language instruction $x$ and visual observation $o_t$, and predicts discrete action tokens $a_{t,1:H}$ that are detokenized into continuous robot control. 
A single target decoding step is
\begin{equation}
  p_\theta(a_i \mid x, o_t, a_{<i}),
  \label{eq:vla_target}
\end{equation}
where $\theta$ denotes the target VLA and $a_{<i}$ are previous action tokens. 
The observation is encoded into visual patch embeddings and combined with text tokens to form the prefill input.  Because closed-loop control repeatedly invokes this decoder over multi-token action horizons, latency scales with the number of target decoding rounds.  Acceleration therefore requires reducing target forward passes while preserving the VLA action-token interface and task success \citep{yang2026efficientvla}.

\subsection{Speculative Decoding and Relaxed Verification}

Speculative decoding accelerates autoregressive generation by using a lightweight draft model to propose several future tokens and a larger target model to verify them in parallel \citep{leviathan2023fast,chen2023accelerating}. Follow-up work improves this paradigm through big--little decoding \citep{kim2023speculative}, tree-based verification \citep{miao2024specinfer}, auxiliary decoding heads \citep{cai2024medusa}, retrieval \citep{he2024rest}, feature-level drafting \citep{li2024eagle}, or self-speculation \citep{zhang2024draft}. These methods reduce the number of target decoding rounds by accepting multiple draft tokens after one target verification pass.

For VLA policies, exact token matching is often overly restrictive because nearby action tokens can detokenize to similar continuous controls. Recent VLA speculative decoding methods therefore introduce relaxed verification \citep{wang2025spec}, with kinematic and hybrid variants adapting acceptance to embodied action spaces \citep{zheng2026kerv,zheng2026heisd}. A common relaxed acceptance rule is
\begin{equation}
  \mathrm{Accept}_\rho(\hat a_i)
  =
  \mathbb 1\!\left[
    d(\hat a_i, a_i^\star) \le \rho
  \right],
  \label{eq:relaxed_acceptance}
\end{equation}
where $d(\cdot,\cdot)$ measures token-space or detokenized action-space distance, $a_i^\star$ is the target token, and $\rho$ is the tolerance. Larger $\rho$ accepts longer draft prefixes and improves speed, while smaller $\rho$ keeps the executed sequence closer to the target decoder.

The limitation is that this tolerance is global, while the physical cost of an accepted deviation is state-dependent. Although $a_i^\star$ is conditioned on the current observation, the accept/reject decision still compresses scene-dependent control risk into a fixed distance threshold. The same draft--target deviation can be harmless in free-space motion but harmful near contact, where manipulation depends on contact configuration and local interaction dynamics \citep{suomalainen2022survey,pang2023global}. A small displacement may change whether the gripper reaches, grasps, collides, or enters the wrong contact phase \citep{dong2021tactile}. Thus, aggressive relaxation improves accepted length but can admit physically unsafe actions, forcing conservative tolerances to preserve task success. WA-SpecDec addresses this bottleneck by improving the shared conditioning state used by proposal and verification, rather than changing the relaxed acceptance rule itself.
\begin{figure*}[t]
  \centering
  \includegraphics[width=\linewidth]{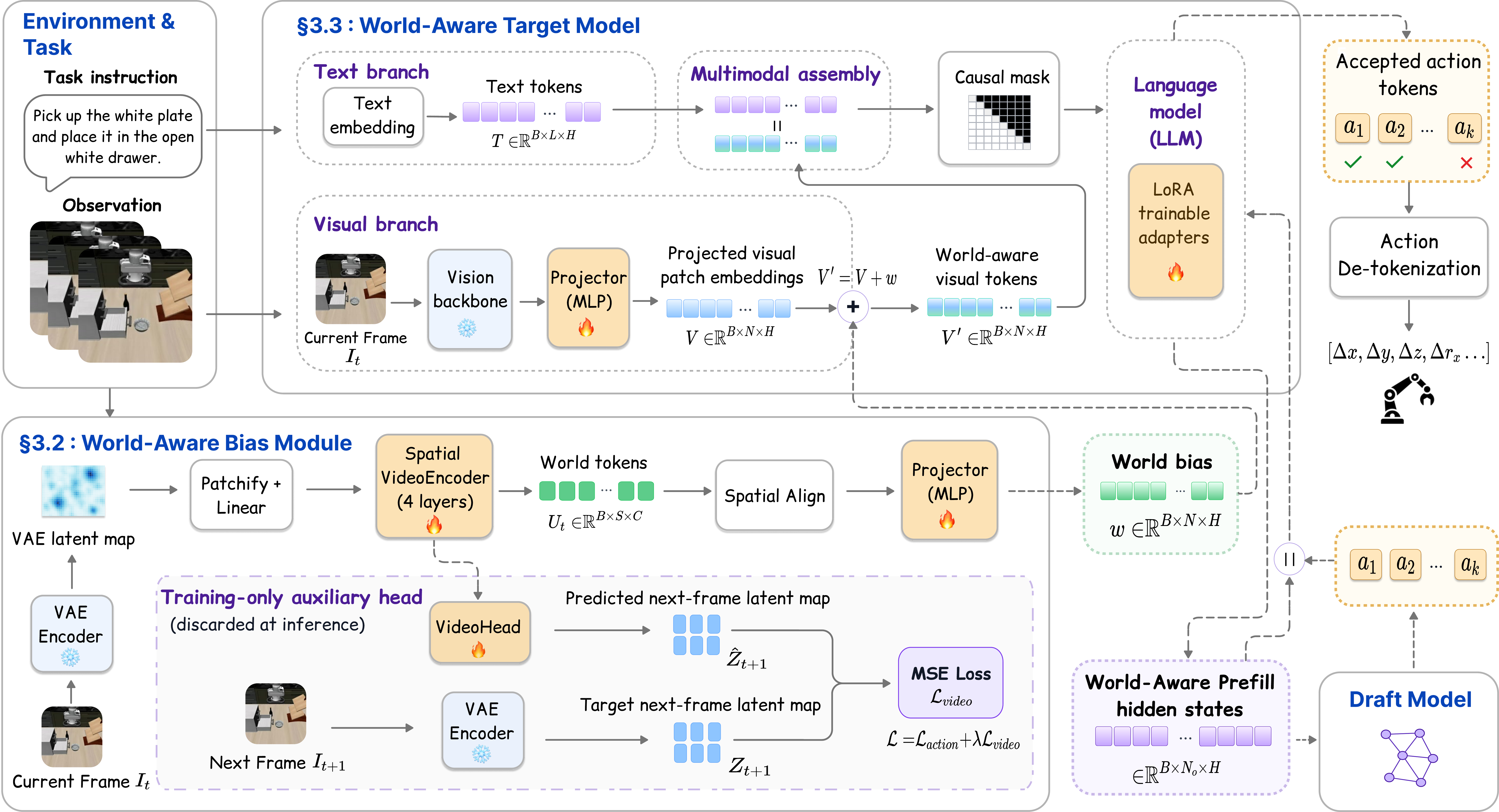}
\caption{
The overall WA-SpecDec architecture. World-aware bias module adds a spatially aligned world bias $w_t$ to visual patch embeddings before prefill computation, yielding world-aware prefill hidden states shared by draft proposal and target verification. The auxiliary next-frame prediction head is used only during training.
}
  \label{fig:wab_structure}
\end{figure*}

\subsection{World Representations for Scene-Aware Decoding}

The reliability of relaxed VLA verification depends not only on the acceptance rule, but also on the scene representation encoded into the shared prefill hidden states. Modern visual backbones provide strong semantic and spatial features \citep{oquab2024dinov,Zhai_2023_ICCV}, yet their pretraining objectives are not designed to preserve control-critical variables such as object pose, gripper--object distance, contact phase, and occlusion \citep{majumdar2023we}. Since token-based VLA policies use visual patch embeddings to form the multimodal prefill input for both drafting and verification, missing physical detail can affect the target reference and draft proposal together \citep{Karamcheti-RSS-23}.

World representations provide a complementary source of predictive physical structure. VAEs and latent world models learn compact representations through reconstruction and future prediction \citep{kingma2014autoencoding,ha2018worldmodels}. Latent dynamics models support predictive control from pixels \citep{hafner2019planet,hansen2024td}, while recent video-predictive representation learning shows that latent compressors and feature-prediction objectives retain cues useful for physical change \citep{bardes2024revisiting,wan2025wan}. WA-SpecDec uses this complementarity to make relaxed speculative decoding more reliable. Instead of adding a post-hoc acceptor, changing $\rho$, or building a larger draft model, it injects a world-model-derived physical-scene bias into visual patch embeddings before prefill computation. The resulting world-aware prefill hidden states are shared by draft proposal and target verification, preserving the original VLA action-token interface and relaxed acceptance rule while giving both sides of speculative decoding access to the same predictive scene signal.

\section{WA-SpecDec Framework}
\label{sec:method}

\subsection{Overview}
\label{sec:method_overview}

WA-SpecDec injects physical-world awareness into relaxed VLA speculative decoding by modifying the visual inputs that produce shared prefill hidden states, rather than changing the acceptance interface. As shown in Figure~\ref{fig:wab_structure}, at each control step the policy receives a language instruction $x$ and current observation $I_t$, and outputs accepted action tokens that are detokenized into continuous robot commands. The speculative decoding loop remains unchanged. A lightweight draft model proposes action tokens, the target VLA verifies them in parallel with a relaxed acceptance rule, and the accepted prefix is executed. The key difference is that both proposal and verification are conditioned on world-aware prefill hidden states instead of hidden states produced from a purely semantic vision-language input.

WA-SpecDec introduces a World-Aware Bias (WAB) module to produce a spatially aligned physical-scene bias $w_t$ from the current observation. This bias is added to the existing visual patch embeddings before prefill computation, so the target-side multimodal sequence length is unchanged and the original VLA action-token interface is preserved. During training, WAB is shaped by the action prediction objective and an auxiliary next-frame latent objective. The auxiliary branch is used only for representation learning and is removed at inference, so deployment requires only the current frame and performs no future-frame prediction.

The resulting world-aware prefill hidden states serve as the common conditioning state for the target and draft models. The target uses these states to define the reference distribution for relaxed verification, while the draft model is distilled to propose tokens from the same states. WA-SpecDec therefore does not introduce a new acceptance rule or a post-hoc acceptor. Instead, it changes the target reference that existing relaxed acceptance rules compare against. This design is intended to improve the acceptance--success frontier and the contact-sensitive reliability measured by near-contact failure. A physically better-conditioned reference can support longer accepted prefixes under looser relaxation while reducing failures that occur after the policy has entered near-contact states. Because WAB adds no extra multimodal tokens, the added computation remains limited to the compact bias path.

Section~\ref{sec:wab} describes how WAB computes the spatially aligned world bias, and Section~\ref{sec:world_aware_prefill} explains how this bias produces the world-aware prefill hidden states and target reference used for drafting and verification.

\subsection{World-Aware Bias Module}
\label{sec:wab}

The World-Aware Bias (WAB) module provides the physical-scene signal used to enrich the visual inputs that produce VLA prefill hidden states. Given the current RGB observation $I_t$, it produces a spatially-aligned bias $w_t$ that has the same shape as the VLA visual patch embeddings. Rather than appending extra world tokens to the multimodal sequence, WAB modulates existing visual tokens before prefill computation, allowing physical information to affect prefill hidden states without increasing the target-side sequence length.

WAB starts from a frozen VAE latent space:
\begin{equation}
  Z_t = E_{\mathrm{vae}}(I_t)
  \in \mathbb{R}^{B \times H_z \times W_z \times C_{\mathrm{vae}}}.
  \label{eq:wab_vae}
\end{equation}
The encoder $E_{\mathrm{vae}}$ maps $I_t$ to a latent scene map $Z_t$, where $B$ is the batch size, $H_z$ and $W_z$ are the latent spatial resolution, and $C_{\mathrm{vae}}$ is the VAE channel dimension. We keep this encoder frozen so that WAB learns only a lightweight adaptation path on top of a stable reconstructive representation. This latent representation is complementary to the semantic vision backbone because it retains spatial appearance and layout details that are useful for contact-sensitive control.

The latent map is then adapted into world tokens:
\begin{equation}
  U_t =
  F_{\omega}\!\left(\phi_{\mathrm{tok}}(Z_t)\right)
  \in \mathbb{R}^{B \times S \times C},
  \qquad S = H_z W_z .
  \label{eq:wab_tokens}
\end{equation}
The tokenization map $\phi_{\mathrm{tok}}$ flattens the VAE spatial grid and applies a linear projection to produce $S$ latent tokens of width $C$. The spatial VideoEncoder $F_{\omega}$ then refines these tokens into $U_t$. This step keeps the VAE latent grid as the source of spatial structure, while allowing the trainable WAB path to emphasize action-relevant factors such as object pose, occlusion, and local gripper--object geometry.

To make these world tokens usable by the VLA visual branch, WAB aligns them to the visual patch grid and projects them to the VLA hidden width:
\begin{equation}
  w_t =
  f_{\mathrm{bias}}\!\left(\mathrm{SpatialAlign}(U_t)\right)
  \in \mathbb{R}^{B \times N_v \times H}.
  \label{eq:wab_bias}
\end{equation}
Here, $N_v$ is the number of VLA visual patches and $H$ is the hidden width of the projected visual embeddings. $\mathrm{SpatialAlign}(\cdot)$ reshapes $U_t$ to its two-dimensional latent grid, interpolates it to the visual patch resolution, and flattens it back to a token sequence. The projector $f_{\mathrm{bias}}$ maps each aligned world token to the VLA hidden width. As a result, $w_t$ can be added directly to the visual patch embeddings $V_t\in\mathbb{R}^{B\times N_v\times H}$, giving each patch a location-specific physical-scene bias.

WAB is trained not only through action prediction but also through a dense next-frame latent prediction objective. From the current-frame world tokens, a training-only VideoHead $g_{\eta}$ predicts the next-frame VAE latent map:
\begin{equation}
  \mathcal{L}_{\mathrm{video}}
  =
  \frac{1}{H_z W_z C_{\mathrm{vae}}}
  \left\|
    g_{\eta}(U_t)
    -
    E_{\mathrm{vae}}(I_{t+1})
  \right\|_2^2 .
  \label{eq:wab_video_loss}
\end{equation}
The next frame $I_{t+1}$ is used only during training to form the target latent map, and $\eta$ denotes the VideoHead parameters. Because the loss is computed over the full latent grid rather than a pooled representation, it encourages $U_t$ to preserve spatially-resolved cues predictive of short-term scene change. This supervision is particularly relevant for relaxed VLA decoding, where small token deviations can become unsafe near contact.

The overall training loss combines the action-token objective with this auxiliary latent prediction term:
\begin{equation}
  \mathcal{L}
  =
  \mathcal{L}_{\mathrm{action}}
  +
  \lambda \mathcal{L}_{\mathrm{video}} .
  \label{eq:wab_total_loss}
\end{equation}
The scalar $\lambda$ controls the strength of the auxiliary objective. The VideoHead is discarded after training, so inference uses only the current observation $I_t$ and does not perform future-frame prediction. The learned WAB path therefore adds only the current-frame bias $w_t$ at test time. This bias is later encoded into the world-aware prefill hidden states shared by both the target and draft, keeping proposal and verification conditioned on the same spatially grounded scene signal.

\subsection{World-Aware Target Model}
\label{sec:world_aware_prefill}

Figure~\ref{fig:wab_structure} shows how the target model becomes world-aware before any action token is decoded. Let the frozen visual backbone and multimodal projector produce visual patch embeddings $V_t \in \mathbb{R}^{B \times N_v \times H}$. WAB produces the spatially-aligned bias $w_t \in \mathbb{R}^{B \times N_v \times H}$ from Equation~\ref{eq:wab_bias}, which matches the shape of $V_t$, and the target injects it as a residual world-conditioned modulation.
\begin{equation}
  V^{\mathrm{wa}}_t = V_t + w_t ,
  \label{eq:wa_visual_tokens}
\end{equation}
The addition is element-wise across visual patches, so each patch receives its own location-specific bias. Although the operation is additive, $w_t$ is not a free parameter because it is computed from VAE latent tokens that retain reconstructive scene detail and is shaped by the dense next-frame latent objective in Equation~\ref{eq:wab_video_loss}. The added direction is therefore trained to summarize scene factors predictive of short-term physical change, such as pose, occlusion, and contact configuration, rather than only image--text semantics. Injecting it before prefill computation makes this physical-scene direction participate in the transformer's query, key, value, and feed-forward projections, so it becomes part of the hidden states and key--value cache used to produce subsequent action-token logits.

The world-aware visual tokens are concatenated with text embeddings $T_x$,
\begin{equation}
  P^{\mathrm{wa}}_t = [T_x; V^{\mathrm{wa}}_t].
  \label{eq:wa_prefill_input}
\end{equation}
The LoRA-adapted target language model converts this input into prefill hidden states,
\begin{equation}
  h^{\mathrm{wa}}_t
  =
  \mathrm{Prefill}_{\theta+\Delta\theta}
  \!\left(P^{\mathrm{wa}}_t\right),
  \label{eq:wa_prefill_states}
\end{equation}
where $\theta$ denotes the pretrained VLA parameters and $\Delta\theta$ denotes the trainable LoRA adapter. This placement is central because the world representation is not used as a post-hoc acceptor, but changes the target reference token itself. Because $w_t$ is added to existing visual tokens rather than appended as extra world tokens, $P^{\mathrm{wa}}_t$ has the same sequence length as the original multimodal prefill input. WA-SpecDec therefore injects a \emph{per-patch physical-scene prior} while preserving the patch-level semantic content of $V_t$ and avoiding additional target self-attention cost.

The action objective in Equation~\ref{eq:wab_total_loss} trains this world-aware target to predict demonstration action tokens. For a training set $\mathcal{D}$ of instruction, observation, and action-token sequences, we use
\begin{equation}
\begin{aligned}
  \mathcal{L}_{\mathrm{action}}
  &=
  -\sum_{(x,I_t,a^\star_{1:H}) \in \mathcal{D}}
  \sum_{i=1}^{H} \\
  &\quad
  \log
  p^{\mathrm{wa}}_{\theta}
  \!\left(
    a^\star_i
    \mid x, I_t, a^\star_{<i}
  \right),
\end{aligned}
  \label{eq:wa_action_loss}
\end{equation}
with
\begin{equation}
  p^{\mathrm{wa}}_{\theta}
  \!\left(a_i \mid x,I_t,a_{<i}\right)
  \triangleq
  p_{\theta+\Delta\theta}
  \!\left(a_i \mid h^{\mathrm{wa}}_t,a_{<i}\right).
  \label{eq:wa_target_distribution}
\end{equation}
Together with $\mathcal{L}_{\mathrm{video}}$, this objective updates WAB and $\Delta\theta$ while keeping the VAE and base visual stack frozen. The VideoHead affects the target only through training-time gradients into WAB. This stage defines the target policy that WA-SpecDec accelerates. We do not claim distribution preservation with respect to the vanilla VLA, but with respect to the world-aware target $p^{\mathrm{wa}}_{\theta}$.

At inference, relaxed verification is unchanged except for the reference distribution. Given a draft token $\hat a_i$, the target produces
\begin{equation}
  a^{\mathrm{wa}}_i
  =
  \arg\max_{a}
  p^{\mathrm{wa}}_{\theta}
  \!\left(a \mid x,I_t,a_{<i}\right),
  \label{eq:wa_reference_token}
\end{equation}
and applies the same distance test as Equation~\ref{eq:relaxed_acceptance},
\begin{equation}
  \mathrm{Accept}^{\mathrm{wa}}_\rho(\hat a_i)
  =
  \mathbb{1}
  \!\left[
    d(\hat a_i,a^{\mathrm{wa}}_i) \le \rho
  \right].
  \label{eq:wa_relaxed_acceptance}
\end{equation}
WA-SpecDec therefore acts below the acceptance policy rather than replacing it. Any relaxed acceptance rule that compares draft proposals with target references can use the world-aware reference. This is where acceleration enters. Because the target and draft share the same physical-scene-aware prefill hidden states, the draft is encouraged to match a better-conditioned reference, yielding longer accepted prefixes at fixed reliability and fewer target verification rounds. Since $w_t$ adds no tokens, this gain is not offset by extra target-side attention cost.

The same world-aware prefill hidden states are used for drafting. After training the world-aware target, we freeze it and distill a lightweight draft model $q_\phi$ against its output distribution,
\begin{equation}
\begin{aligned}
  \mathcal{L}_{\mathrm{draft}}
  =
  \mathbb{E}_{(x,I_t,a_{<i})}
  \Big[
  \mathrm{KL}\big(
    p^{\mathrm{wa}}_{\theta}
    (\cdot \mid x,I_t,a_{<i})
    \,\|\,
  \\
    q_\phi(\cdot \mid h^{\mathrm{wa}}_t,a_{<i})
  \big)
  \Big].
\end{aligned}
\label{eq:wa_draft_loss}
\end{equation}
Sharing $h^{\mathrm{wa}}_t$ couples proposal and verification under the same physical-scene conditioning. The draft is not strengthened in isolation. It proposes tokens from the same world-aware prefill hidden states that also define the target reference. In deployment, WAB computes $w_t$ once per observation. The draft model proposes $\gamma$ tokens from $h^{\mathrm{wa}}_t$, and the target verifies them in parallel using Equation~\ref{eq:wa_relaxed_acceptance}. The accepted prefix is executed, while the first rejected draft token is replaced by the corresponding target token before the next speculative round.
\begin{table*}[t]
  \centering
  \small
  \setlength{\tabcolsep}{4.2pt}
  \renewcommand{\arraystretch}{1.12}
  \begin{tabular*}{\textwidth}{@{\extracolsep{\fill}}ll*{4}{ccc}@{}}
    \toprule
    \multicolumn{2}{c}{Decoding Method}
    & \multicolumn{3}{c}{Spatial}
    & \multicolumn{3}{c}{Object}
    & \multicolumn{3}{c}{Goal}
    & \multicolumn{3}{c}{Long} \\
    \cmidrule(lr){1-2}
    \cmidrule(lr){3-5}
    \cmidrule(lr){6-8}
    \cmidrule(lr){9-11}
    \cmidrule(l){12-14}
    Relaxation & Policy
    & Succ. & Spd. & NCF
    & Succ. & Spd. & NCF
    & Succ. & Spd. & NCF
    & Succ. & Spd. & NCF \\
    \midrule
    -- & AR: OpenVLA
    & 82.8 & $1.00\times$ & 73.3
    & 69.8 & $1.00\times$ & 49.7
    & 78.2 & $1.00\times$ & 46.8
    & 57.8 & $1.00\times$ & 45.5 \\
    \midrule
    $\rho$ & Standard SpecDec
    & 79.4 & $1.26\times$ & 78.6
    & 63.0 & $1.13\times$ & 60.0
    & 69.6 & $1.23\times$ & 64.5
    & 49.4 & $1.18\times$ & 52.2 \\
    $\rho$ & WA-SpecDec
    & 83.0 & $1.54\times$ & 72.9
    & 68.8 & $1.24\times$ & 48.7
    & 75.8 & $1.32\times$ & 46.3
    & 56.2 & $1.21\times$ & 44.7 \\
    \midrule

    KERV & Standard SpecDec
    & 80.2 & $1.52\times$ & 75.8
    & 63.4 & $1.54\times$ & 63.9
    & 70.8 & $1.61\times$ & 61.6
    & 48.6 & $1.41\times$ & 54.1 \\
    KERV & WA-SpecDec
    & 83.4 & $1.59\times$ & 72.3
    & 68.4 & $1.64\times$ & 48.1
    & 77.6 & $1.67\times$ & 47.3
    & 52.8 & $1.45\times$ & 42.8 \\
    \midrule

    HeiSD & Standard SpecDec
    & 77.4 & $1.74\times$ & 69.0
    & 68.0 & $2.26\times$ & 63.8
    & 70.8 & $2.10\times$ & 66.4
    & 50.4 & $1.71\times$ & 51.6 \\
    HeiSD & WA-SpecDec
    & 80.4 & $1.77\times$ & 65.3
    & 70.6 & $2.31\times$ & 49.0
    & 78.0 & $2.12\times$ & 43.6
    & 57.0 & $1.79\times$ & 42.3 \\
    \bottomrule
  \end{tabular*}
\caption{
Main comparison on four LIBERO task suites under identical relaxation schemes.  \textbf{Succ.} denotes task success rate (\%), \textbf{Spd.} denotes wall-clock speedup over the AR target, and \textbf{NCF} follows the definition in Section~\ref{sec:exp_setup}. 
Speedup includes the WAB overhead for WA-SpecDec.
}
  \label{tab:matched_speedup}
\end{table*}

\begin{table}[t]
\centering
\small
\begin{tabular}{llccc}
\toprule
Relax. & Policy & Succ. & Spd. & NCF \\
\midrule
-- 
& AR ActionCodec 
& 95.8 
& $1.00\times$ 
& 76.2 \\
\midrule

$\rho$
& Standard SpecDec
& 88.6
& $1.18\times$
& 88.7 \\

& WA-SpecDec
& 95.2
& $1.25\times$
& 70.8 \\
\midrule

KERV
& Standard SpecDec
& 91.2
& $1.57\times$
& 84.1 \\

& WA-SpecDec
& 95.6
& $1.72\times$
& 77.3 \\
\midrule

HeiSD
& Standard SpecDec
& 88.0
& $2.41\times$
& 86.7 \\

& WA-SpecDec
& 94.2
& $2.63\times$
& 72.4 \\
\bottomrule
\end{tabular}
\caption{
Performance of the stronger ActionCodec target policy on LIBERO-Goal.
Standard SpecDec and WA-SpecDec use the same target policy and decoding configuration under each relaxation scheme. Relax. denotes the relaxation scheme.
}
\label{tab:actioncodec}
\end{table}

\section{Experiments}
\label{sec:experiments}

\subsection{Experimental Setup}
\label{sec:exp_setup}

\paragraph{Target policies.}
We evaluate WA-SpecDec across multiple target policies and evaluation settings.
Our primary LIBERO experiments use a fine-tuned \textbf{OpenVLA} policy \citep{pmlr-v270-kim25c}.
To assess whether the effectiveness of WA-SpecDec extends to stronger target policies, we additionally consider an \textbf{ActionCodec} \citep{dong2026actioncodecmakesgoodaction} target on LIBERO and \textbf{UniVLA} \citep{bu2025univla} on SIMPLER-Env \citep{li24simpler}.
We further evaluate OpenVLA on SIMPLER-Env, allowing us to separately assess generalization across environments and across target-policy strengths.
For each setting, the autoregressive variant of the corresponding target policy, denoted \textbf{AR target}, serves as the reliability reference.

\paragraph{Baselines and verifiers.}
We compare \textbf{WA-SpecDec} against \textbf{Standard SpecDec}, namely standard speculative decoding without world-aware conditioning, corresponding to the Spec-VLA baseline \citep{wang2025spec}, under the same target policy and decoding configuration.
We consider three relaxed verification schemes: the $\rho$-based token-distance verifier \citep{wang2025spec}, \textbf{KERV} \citep{zheng2026kerv}, and \textbf{HeiSD} \citep{zheng2026heisd}.
KERV incorporates kinematic information into verification, whereas HeiSD combines token- and kinematic-level criteria.
Across all settings, the target model, verifier, and decoding configuration are held fixed between Standard SpecDec and WA-SpecDec, thereby isolating the contribution of world-aware conditioning.

\paragraph{Benchmarks and task suites.}
We evaluate WA-SpecDec on two manipulation benchmarks.
Our primary evaluation is conducted on \textbf{LIBERO} \citep{liu2023libero}, covering LIBERO-Spatial, LIBERO-Object, LIBERO-Goal, and LIBERO-Long.
These suites span diverse manipulation behaviors, including reaching, grasping, placement, object relocation, and long-horizon task execution, with varying degrees of contact sensitivity.
To assess generalization beyond LIBERO, we additionally evaluate on \textbf{SIMPLER-Env} \citep{li24simpler} using four manipulation tasks: Carrot, Stack, Spoon, and Eggplant.

\paragraph{Metrics.}
\textbf{Task success} is the percentage of evaluation episodes completing the language-specified manipulation task. \textbf{Accepted length} is the average number of draft action tokens accepted per target verification step.
\textbf{Wall-clock speedup} is measured relative to the AR target and includes draft proposal, target verification, WAB computation, and cache updates. \textbf{NCF} (Near-contact failure) measures contact-sensitive failure among unsuccessful rollouts. For each failed episode, we compute the minimum simulator-state distance from the gripper fingertip sites to either the task-relevant object or the target receptacle across timesteps. NCF is the percentage of failed episodes whose minimum distance falls below a threshold \(\delta\) before termination.


\paragraph{Implementation details.}
Appendix~\ref{app:implementation} provides implementation details. Code and evaluation scripts will be released upon acceptance.


\begin{figure}[t]
  \centering
  \includegraphics[width=\columnwidth]{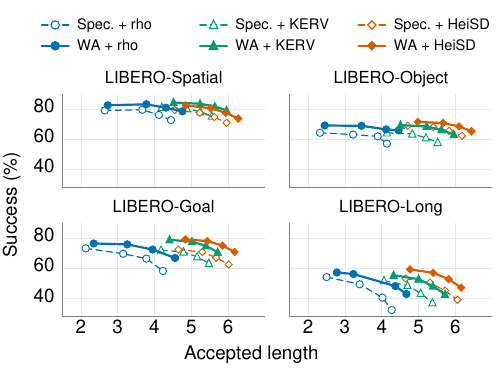}
\caption{
Accepted-length--success frontiers on four LIBERO task suites. Each curve sweeps relaxation within a fixed relaxed acceptance rule. Solid lines denote WA-SpecDec and dashed lines denote Standard SpecDec.
}
  \label{fig:accept_success}
\end{figure}

\subsection{Main Results on LIBERO}
\label{sec:exp_main}

Table~\ref{tab:matched_speedup} presents the main results with OpenVLA under three relaxed verifier families, including $\rho$, KERV, and HeiSD. Across all verifier families and task suites, WA SpecDec consistently achieves a better tradeoff between reliability and acceleration than Standard SpecDec, remaining closer to the AR target while retaining net wall clock acceleration. Under the $\rho$ verifier, WA SpecDec improves success from $63.0$ to $68.8$ on Object, from $69.6$ to $75.8$ on Goal, and from $49.4$ to $56.2$ on Long, while remaining comparable to the AR target on Spatial. These gains are obtained without sacrificing speed. For example, speedup increases from $1.26\times$ to $1.54\times$ on Spatial and from $1.23\times$ to $1.32\times$ on Goal. The same pattern holds with stronger relaxed verifiers. With HeiSD, WA SpecDec improves success from $70.8$ to $78.0$ on Goal and from $50.4$ to $57.0$ on Long, while also achieving comparable or higher speedup. Together, these results show that world aware conditioning complements existing relaxed verification rules and remains effective across different verifier designs.

Importantly, the gains are not specific to the OpenVLA target. Table~\ref{tab:actioncodec} evaluates the same decoding framework with the stronger ActionCodec target on LIBERO Goal, whose AR policy achieves $95.8\%$ success. Across all three relaxation schemes, WA SpecDec preserves substantially more of the AR target performance than Standard SpecDec while also achieving higher speedup. Under $\rho$, success increases from $88.6\%$ to $95.2\%$ while speedup improves from $1.18\times$ to $1.25\times$. With KERV, WA SpecDec reaches $95.6\%$ success, nearly matching the $95.8\%$ AR target, while accelerating inference by $1.72\times$, compared with $91.2\%$ success and $1.57\times$ speedup for Standard SpecDec. Under HeiSD, WA SpecDec achieves $94.2\%$ success at $2.63\times$ speedup, whereas Standard SpecDec achieves $88.0\%$ success at $2.41\times$. These results demonstrate that the benefit of world aware conditioning persists with a substantially stronger target policy and is therefore compatible with improvements to the underlying VLA action representation.

The NCF results further support the contact sensitive motivation. Across all Standard SpecDec and WA SpecDec pairs in Table~\ref{tab:matched_speedup}, WA SpecDec reduces NCF by $18.6\%$ on average relative to the corresponding speculative baselines. The same pattern is observed with ActionCodec, where NCF decreases from $88.7$ to $70.8$ under $\rho$, from $84.1$ to $77.3$ under KERV, and from $86.7$ to $72.4$ under HeiSD. Since NCF measures failures after the policy has entered the near contact region, these reductions indicate that the success improvements are concentrated in states where small accepted action deviations are more likely to cause collision, missed grasp, or placement failure. This suggests that WA SpecDec improves not only decoding efficiency but also the reliability of accepted actions in contact sensitive states.

Figure~\ref{fig:accept_success} provides a complementary view through the full acceptance and success frontier over a wider relaxation sweep. Within each verifier family, the WA SpecDec curves generally shift upward and to the right relative to Standard SpecDec. The upward shift indicates higher task success at comparable accepted length, while the rightward shift indicates longer accepted prefixes at comparable success. Because longer accepted prefixes reduce the number of target verification rounds per action sequence, this frontier shift provides further evidence that WA SpecDec enables more reliable acceleration rather than simply accepting more permissive draft predictions.

\begin{table*}[t]
  \centering
  \small
  \setlength{\tabcolsep}{4.5pt}
  \renewcommand{\arraystretch}{1.12}
  \begin{tabular*}{\textwidth}{@{\extracolsep{\fill}}ll*{4}{cc}@{}}
    \toprule
    \multicolumn{2}{c}{Decoding Method}
    & \multicolumn{2}{c}{Carrot}
    & \multicolumn{2}{c}{Stack}
    & \multicolumn{2}{c}{Spoon}
    & \multicolumn{2}{c}{Eggplant} \\
    \cmidrule(lr){1-2}
    \cmidrule(lr){3-4}
    \cmidrule(lr){5-6}
    \cmidrule(lr){7-8}
    \cmidrule(l){9-10}
    Relaxation & Policy
    & Succ. & Spd.
    & Succ. & Spd.
    & Succ. & Spd.
    & Succ. & Spd. \\
    \midrule
    -- & AR: OpenVLA
    & 25.00 & $1.00\times$
    & 12.50 & $1.00\times$
    & 33.33 & $1.00\times$
    & 16.67 & $1.00\times$ \\
    \midrule

    $\rho$ & Standard SpecDec
    & 16.67 & $1.31\times$
    & 8.33 & $1.27\times$
    & 20.83 & $1.14\times$
    & 12.50 & $1.28\times$ \\

    $\rho$ & WA-SpecDec
    & 29.17 & $1.53\times$
    & 12.50 & $1.34\times$
    & 33.33 & $1.43\times$
    & 20.83 & $1.54\times$ \\
    \midrule

    KERV & Standard SpecDec
    & 12.50 & $1.41\times$
    & 8.33 & $1.52\times$
    & 16.67 & $1.53\times$
    & 12.50 & $1.66\times$ \\

    KERV & WA-SpecDec
    & 25.00 & $1.78\times$
    & 16.67 & $1.67\times$
    & 41.67 & $1.72\times$
    & 25.00 & $1.87\times$ \\
    \midrule

    HeiSD & Standard SpecDec
    & 25.00 & $2.16\times$
    & 12.50 & $1.80\times$
    & 29.17 & $2.13\times$
    & 12.50 & $2.14\times$ \\

    HeiSD & WA-SpecDec
    & 33.33 & $2.98\times$
    & 20.83 & $2.42\times$
    & 41.67 & $2.60\times$
    & 25.00 & $2.56\times$ \\
    \bottomrule
  \end{tabular*}
  \caption{
  Performance with the OpenVLA target policy on four SIMPLER Env tasks. 
  \textbf{Succ.} denotes task success rate (\%) and \textbf{Spd.} denotes wall clock speedup over the corresponding AR target. 
  Standard SpecDec and WA-SpecDec use the same target policy and decoding configuration under each relaxation scheme.
  }
  \label{tab:simpler_openvla}
\end{table*}

\begin{table*}[t]
  \centering
  \small
  \setlength{\tabcolsep}{4.5pt}
  \renewcommand{\arraystretch}{1.12}
  \begin{tabular*}{\textwidth}{@{\extracolsep{\fill}}ll*{4}{cc}@{}}
    \toprule
    \multicolumn{2}{c}{Decoding Method}
    & \multicolumn{2}{c}{Carrot}
    & \multicolumn{2}{c}{Stack}
    & \multicolumn{2}{c}{Spoon}
    & \multicolumn{2}{c}{Eggplant} \\
    \cmidrule(lr){1-2}
    \cmidrule(lr){3-4}
    \cmidrule(lr){5-6}
    \cmidrule(lr){7-8}
    \cmidrule(l){9-10}
    Relaxation & Policy
    & Succ. & Spd.
    & Succ. & Spd.
    & Succ. & Spd.
    & Succ. & Spd. \\
    \midrule
    -- & AR: UniVLA
    & 0.625 & $1.00\times$
    & 0.208 & $1.00\times$
    & 0.583 & $1.00\times$
    & 0.833 & $1.00\times$ \\
    \midrule

    $\rho$ & Standard SpecDec
    & 0.417 & $1.26\times$
    & 0.042 & $1.19\times$
    & 0.375 & $1.23\times$
    & 0.708 & $1.20\times$ \\

    $\rho$ & WA-SpecDec
    & 0.583 & $1.43\times$
    & 0.208 & $1.56\times$
    & 0.625 & $1.38\times$
    & 0.875 & $1.48\times$ \\
    \midrule

    KERV & Standard SpecDec
    & 0.375 & $1.44\times$
    & 0.125 & $1.50\times$
    & 0.375 & $1.43\times$
    & 0.667 & $1.71\times$ \\

    KERV & WA-SpecDec
    & 0.625 & $1.67\times$
    & 0.208 & $1.74\times$
    & 0.583 & $1.78\times$
    & 0.833 & $1.92\times$ \\
    \midrule

    HeiSD & Standard SpecDec
    & 0.542 & $2.31\times$
    & 0.083 & $2.04\times$
    & 0.583 & $1.95\times$
    & 0.750 & $2.23\times$ \\

    HeiSD & WA-SpecDec
    & 0.708 & $2.78\times$
    & 0.250 & $2.55\times$
    & 0.625 & $2.67\times$
    & 0.917 & $2.36\times$ \\
    \bottomrule
  \end{tabular*}
  \caption{
  Performance with the stronger UniVLA target policy on four SIMPLER Env tasks. 
  \textbf{Succ.} denotes task success rate and \textbf{Spd.} denotes wall clock speedup over the corresponding AR target. 
  Standard SpecDec and WA-SpecDec use the same target policy and decoding configuration under each relaxation scheme.
  }
  \label{tab:simpler_univla}
\end{table*}

\subsection{Generalization to SIMPLER Env}
\label{sec:simpler}

To evaluate whether the effectiveness of WA SpecDec extends beyond LIBERO, we further conduct experiments on four tasks in SIMPLER Env using both OpenVLA and the stronger UniVLA target policy. Tables~\ref{tab:simpler_openvla} and~\ref{tab:simpler_univla} report results under the same three relaxation schemes used in our LIBERO evaluation. Across both target policies, all four tasks, and all three relaxation schemes, WA SpecDec consistently improves both task success and wall clock speedup over Standard SpecDec. These results show that the benefit of world aware conditioning generalizes beyond LIBERO and remains effective across different target policies.

Table~\ref{tab:simpler_openvla} reports the results with OpenVLA. Under $\rho$, WA SpecDec improves average success across the four tasks from $14.6\%$ to $24.0\%$ while increasing average speedup from $1.25\times$ to $1.46\times$. With KERV, average success increases from $12.5\%$ to $27.1\%$, together with an improvement in speedup from $1.53\times$ to $1.76\times$. The same trend holds under HeiSD, where WA SpecDec achieves $30.2\%$ average success at $2.64\times$ speedup, compared with $19.8\%$ success at $2.06\times$ for Standard SpecDec. The consistent gains across all four tasks indicate that world aware conditioning improves the reliability and acceleration tradeoff under different relaxed verification rules in the SIMPLER Env setting.

Table~\ref{tab:simpler_univla} further evaluates WA SpecDec with the stronger UniVLA target policy. Under $\rho$, average success improves from $0.386$ to $0.573$ while average speedup increases from $1.22\times$ to $1.46\times$. With KERV, average success improves from $0.386$ to $0.562$ and speedup from $1.52\times$ to $1.78\times$. Under HeiSD, WA SpecDec achieves an average success of $0.625$ at $2.59\times$ speedup, compared with $0.490$ at $2.13\times$ for Standard SpecDec. WA SpecDec therefore retains its advantage when paired with a stronger VLA policy, supporting its compatibility with improvements in the underlying target model as well as its generalization to a different manipulation benchmark.

\begin{table*}[t]
  \centering
  \small
  \setlength{\tabcolsep}{2.2pt}
  \renewcommand{\arraystretch}{1.05}
  \begin{tabular}{lccccccccc}
    \toprule
    & -- & \multicolumn{2}{c}{\(\rho=5\)} & \multicolumn{2}{c}{\(\rho=10\)} & \multicolumn{2}{c}{\(\rho=15\)} & \multicolumn{2}{c}{\(\rho=20\)} \\
    \cmidrule(lr){2-2}\cmidrule(lr){3-4}\cmidrule(lr){5-6}\cmidrule(lr){7-8}\cmidrule(lr){9-10}
    Suite & AR target & Spec-VLA & \textbf{WA-SpecDec} & Spec-VLA & \textbf{WA-SpecDec} & Spec-VLA & \textbf{WA-SpecDec} & Spec-VLA & \textbf{WA-SpecDec} \\
    \midrule
    Spatial & 82.8\% & 79.0\% & \textbf{82.4\%} & 79.4\% & \textbf{83.0\%} & 76.0\% & \textbf{80.8\%} & 72.6\% & \textbf{78.4\%} \\
    Object  & 69.8\% & 64.2\% & \textbf{69.0\%} & 63.0\% & \textbf{68.8\%} & 61.8\% & \textbf{66.4\%} & 57.0\% & \textbf{65.8\%} \\
    Goal    & 78.2\% & 73.2\% & \textbf{76.4\%} & 69.6\% & \textbf{75.8\%} & 66.4\% & \textbf{72.4\%} & 58.2\% & \textbf{66.8\%} \\
    Long    & 57.8\% & 54.2\% & \textbf{57.2\%} & 49.4\% & \textbf{56.2\%} & 40.6\% & \textbf{48.2\%} & 32.4\% & \textbf{43.0\%} \\
    \midrule
    Mean    & 72.2\% & 67.7\% & \textbf{71.3\%} & 65.4\% & \textbf{71.0\%} & 61.2\% & \textbf{67.0\%} & 55.1\% & \textbf{63.5\%} \\
    \bottomrule
  \end{tabular}
\caption{
Effect of progressively increasing the relaxation tolerance $\rho$ on task success across four LIBERO task suites.
Spec-VLA and WA-SpecDec are compared under the same $\rho$-based relaxed acceptance rule at each tolerance value.}
  \label{tab:fixed_rho}
\end{table*}

\begin{table}[t]
  \centering
  \small
  \begin{tabular}{@{}lccc@{}}
    \toprule
    Method & Success & Accepted Len. & Speedup \\
    \midrule
    AR target & 72.2\% & -- & \(1.00\times\) \\
    WA target-only & 73.5\% & -- & \(0.97\times\) \\
    Spec-VLA & 65.4\% & 3.59 & \(1.20\times\) \\
    WA w/o video loss & 69.4\% & 4.09 & \(1.31\times\) \\
    \textbf{WA-SpecDec full} & 71.0\% & 4.22 & \(1.34\times\) \\
    \bottomrule
  \end{tabular}
  \caption{
Component ablation averaged over four LIBERO task suites. Speculative variants use $\rho=10$. ``--`` denotes N/A for accepted length.}
  \label{tab:ablation}
\end{table}

\subsection{Robustness to Relaxed Tolerance}
\label{sec:exp_rho_robustness}

Table~\ref{tab:fixed_rho} isolates the effect of the $\rho$-based relaxed acceptance rule by comparing Spec-VLA and WA-SpecDec under the same tolerance values. As $\rho$ increases, Spec-VLA degrades sharply, with mean success dropping from $67.7$ at $\rho=5$ to $55.1$ at $\rho=20$. WA-SpecDec degrades more gracefully, maintaining mean success of $71.3$, $71.0$, $67.0$, and $63.5$ from $\rho=5$ to $\rho=20$. This fixed-rule comparison shows that the gain does not come from changing the acceptance rule, but from making the same permissive rule less damaging through shared world-aware prefill hidden states and a more physically informed target reference.



\subsection{Robustness to the NCF Threshold}
\label{sec:ncf_threshold}

Our default NCF threshold of $2$ cm is physically motivated by the approximate width of one gripper finger in the LIBERO simulator, providing a meaningful scale for identifying failures that occur after the policy enters a contact sensitive region. This threshold is fixed independently of the observed improvement and is not selected to maximize the difference between WA SpecDec and Standard SpecDec. To evaluate whether the NCF conclusion depends on this particular choice, we additionally measure NCF using thresholds of $1$, $2$, $3$, and $4$ cm across all four LIBERO task suites.

Table~\ref{tab:ncf_threshold} shows that WA SpecDec achieves lower NCF than Standard SpecDec under every relaxation scheme and at every evaluated threshold. Averaged across $\rho$, KERV, and HeiSD, WA SpecDec reduces NCF by $12.54$, $11.77$, $10.87$, and $8.03$ percentage points at thresholds of $1$, $2$, $3$, and $4$ cm, respectively. The improvement therefore persists over a broad range of definitions of the near contact region rather than depending on the default $2$ cm threshold.

The reduction is largest at the tighter $1$ cm and $2$ cm thresholds, where contact geometry is most critical to manipulation success. This pattern is consistent with the motivation of WA SpecDec, since small accepted action deviations become increasingly consequential as the gripper approaches the object or receptacle. These results show that the NCF conclusion is robust to the threshold choice, while the default $2$ cm threshold retains a direct physical interpretation in the LIBERO environment.

\subsection{Ablation Study}
\label{sec:exp_ablation}

Table~\ref{tab:ablation} ablates the main components of WA-SpecDec at $\rho=10$. WA target-only adds WAB to the autoregressive target without speculative drafting or relaxed acceptance, improving reliability but running slightly slower due to WAB overhead. This shows that physical-scene conditioning improves the target reference but does not accelerate decoding alone. Spec-VLA improves speed but incurs a large success drop. Adding WAB without the auxiliary video loss partly recovers this gap, while the full model achieves the best accepted length and speedup with success closest to the AR target. These results show that both shared world-aware prefill states and predictive latent supervision are needed for the acceleration--reliability trade-off.


\begin{table}[t]
\centering
\footnotesize
\setlength{\tabcolsep}{2.5pt}
\renewcommand{\arraystretch}{1.12}
\begin{tabular}{@{}llcccc@{}}
\toprule
Relax. & Policy & 1 cm & 2 cm & 3 cm & 4 cm \\
\midrule
-- & AR
& 36.83 & 51.17 & 59.72 & 62.87 \\
\midrule

$\rho$ & Standard SpecDec
& 44.64 & 60.89 & 67.93 & 72.35 \\
$\rho$ & WA SpecDec
& 35.20 & 50.26 & 58.07 & 63.12 \\
\midrule

KERV & Standard SpecDec
& 46.85 & 61.46 & 68.15 & 73.34 \\
KERV & WA SpecDec
& 32.82 & 49.24 & 56.75 & 62.78 \\
\midrule

HeiSD & Standard SpecDec
& 45.31 & 60.72 & 67.43 & 70.21 \\
HeiSD & WA SpecDec
& 31.17 & 48.25 & 56.07 & 65.90 \\
\bottomrule
\end{tabular}

\caption{
Sensitivity of NCF to the near contact distance threshold on LIBERO. Values are averaged across the four task suites.}
\label{tab:ncf_threshold}
\end{table}
\section{Conclusion}
\label{sec:conclusion}
In this study, we presented WA-SpecDec, a world-aware speculative decoding framework for VLA models. By injecting world-model-derived physical scene awareness before prefill computation, WA-SpecDec conditions draft proposal and target verification on shared world-aware prefill hidden states, improving relaxed decoding without changing the acceptance rule. Experiments on four LIBERO task suites show that WA-SpecDec achieves a better reliability--acceleration trade-off, preserving higher success under looser relaxation, enabling longer accepted prefixes, and reducing near-contact failure while retaining speedup. These results suggest that physically informed shared conditioning is an effective approach for improving the reliability of accelerated VLA action decoding.
\section*{Ethics Statement}

This work evaluates efficient VLA decoding in simulated manipulation environments using publicly available research artifacts, including the LIBERO benchmark and existing VLA/model components. It does not involve human subjects, personal data, or real-world robot deployment. All external models, datasets, and simulators are used only for research evaluation, and any released code and evaluation scripts will follow their licenses and terms of use. Future real-world use should follow standard robot deployment practices, including task-specific validation, appropriate supervision, safety checks, and compliance with the licenses of the underlying models, datasets, and simulators.
\bibliography{custom}
\appendix
\begin{algorithm*}[t]
\small
\caption{WA-SpecDec training and inference}
\label{alg:waspecdec}
\begin{algorithmic}[1]
\Require Training set $\mathcal{D}=\{(x,I_t,I_{t+1},a^\star_{1:H})\}$, where $x$ is the instruction, $I_t$ the current image, $I_{t+1}$ the training-only next image, $a^\star_{1:H}$ the demonstration action tokens, and $H$ the number of action tokens decoded for one control query. Frozen VAE encoder $E_{\mathrm{vae}}$, visual embeddings $V_t$, text embeddings $T_x$, and frozen base VLA parameters $\theta$. Trainable modules: token projector $\phi_{\mathrm{tok}}$, spatial VideoEncoder $F_{\omega}$ over VAE latent tokens, bias projector $f_{\mathrm{bias}}$, training-only latent prediction head $g_{\eta}$, and LoRA adapter $\Delta\theta$. Auxiliary weight $\lambda$. Draft model $q_\phi$, proposal operator $\mathrm{Draft}_{\phi}$, block length $\gamma$, distance $d$, tolerance $\rho$, candidate action token $a$, and detokenizer $\mathrm{Detok}$.
\Ensure Trained target $p^{\mathrm{wa}}_{\theta}$, trained draft $q_\phi$, and continuous control sequence $u_{1:H}$ for a current query $(x,I_t)$.
\Statex \textbf{Train the world-aware target}
\For{each minibatch of samples $(x,I_t,I_{t+1},a^\star_{1:H})$ from $\mathcal{D}$}
  \State $Z_t \gets E_{\mathrm{vae}}(I_t)$ \Comment{Eq.~\ref{eq:wab_vae}}
  \State $U_t \gets F_{\omega}\!\left(\phi_{\mathrm{tok}}(Z_t)\right)$ \Comment{Eq.~\ref{eq:wab_tokens}}
  \State $w_t \gets f_{\mathrm{bias}}\!\left(\mathrm{SpatialAlign}(U_t)\right)$ \Comment{Eq.~\ref{eq:wab_bias}}
  \State $V^{\mathrm{wa}}_t \gets V_t + w_t$ and $P^{\mathrm{wa}}_t \gets [T_x,V^{\mathrm{wa}}_t]$ \Comment{Eqs.~\ref{eq:wa_visual_tokens}--\ref{eq:wa_prefill_input}}
  \State $h^{\mathrm{wa}}_t \gets \mathrm{Prefill}_{\theta+\Delta\theta}(P^{\mathrm{wa}}_t)$ \Comment{Eq.~\ref{eq:wa_prefill_states}}
  \State Compute $\mathcal{L}_{\mathrm{action}}$ from $p^{\mathrm{wa}}_{\theta}$ and $a^\star_{1:H}$. \Comment{Eq.~\ref{eq:wa_action_loss}}
  \State Compute $\mathcal{L}_{\mathrm{video}}$ from $g_{\eta}(U_t)$ and $E_{\mathrm{vae}}(I_{t+1})$. \Comment{Eq.~\ref{eq:wab_video_loss}}
  \State $\mathcal{L} \gets \mathcal{L}_{\mathrm{action}}+\lambda\mathcal{L}_{\mathrm{video}}$ \Comment{Eq.~\ref{eq:wab_total_loss}}
  \State Update $\phi_{\mathrm{tok}}$, $F_{\omega}$, $f_{\mathrm{bias}}$, $g_{\eta}$, and $\Delta\theta$ by minimizing $\mathcal{L}$, keeping the frozen modules fixed.
\EndFor
\State Freeze $p^{\mathrm{wa}}_{\theta}$ and train $q_\phi$ by minimizing $\mathcal{L}_{\mathrm{draft}}$. \Comment{Eqs.~\ref{eq:wa_target_distribution}, \ref{eq:wa_draft_loss}}
\Statex \textbf{Run one inference control query}
\State Discard $g_{\eta}$ and do not use $I_{t+1}$ or $\mathcal{L}_{\mathrm{video}}$ at inference.
\State Compute $Z_t$, $U_t$, $w_t$, $V^{\mathrm{wa}}_t$, $P^{\mathrm{wa}}_t$, and $h^{\mathrm{wa}}_t$ from the current $(x,I_t)$. \Comment{Eqs.~\ref{eq:wab_vae}, \ref{eq:wab_tokens}, \ref{eq:wab_bias}, \ref{eq:wa_visual_tokens}--\ref{eq:wa_prefill_states}}
\State $A \gets ()$ and $i \gets 1$
\While{$i \le H$}
  \State $r \gets \min(\gamma,H-i+1)$ and $J \gets \{i,\ldots,i+r-1\}$
  \State $\hat a_{i:i+r-1} \gets \mathrm{Draft}_{\phi}(h^{\mathrm{wa}}_t,A,r)$
  \State In one target verification pass, compute $a^{\mathrm{wa}}_j \gets \arg\max_a p^{\mathrm{wa}}_{\theta}(a\mid h^{\mathrm{wa}}_t,a_{<j})$ for all $j\in J$. \Comment{Eq.~\ref{eq:wa_reference_token}}
  \State Use prefix $a_{<j}=(A,\hat a_{i:j-1})$ for each verified position $j$.
  \State $m \gets 0$
  \While{$m<r$ \textbf{and} $d(\hat a_{i+m},a^{\mathrm{wa}}_{i+m})\le\rho$} \Comment{Eq.~\ref{eq:wa_relaxed_acceptance}}
    \State Append $\hat a_{i+m}$ to $A$ and set $m \gets m+1$
  \EndWhile
  \If{$m<r$}
    \State Append $a^{\mathrm{wa}}_{i+m}$ to $A$ \Comment{replace the first rejected draft token}
    \State $i \gets i+m+1$
  \Else
    \State $i \gets i+m$
  \EndIf
\EndWhile
\State $u_{1:H} \gets \mathrm{Detok}(A)$
\State \Return $u_{1:H}$
\end{algorithmic}
\end{algorithm*}

\section{Implementation details}
\label{app:implementation}
We use the same fine-tuned OpenVLA target policy across all methods. 
The evaluation covers four LIBERO task suites: LIBERO-Spatial, LIBERO-Object, LIBERO-Goal, and LIBERO-Long. For evaluation, we run 50 rollouts per task, resulting in 500 evaluation episodes per suite. 
All methods are evaluated under the same task splits, rollout budget, and simulator settings.

For NCF computation, we define a near-contact state using simulator-state geometry. 
For each failed episode, we compute the minimum distance from the gripper fingertip sites to either the task-relevant object or the target receptacle over all timesteps. 
The episode is counted as a near-contact failure if this minimum distance falls below a fixed threshold $\delta=2\,\mathrm{cm}$ before termination. 
The same threshold is used across all methods, task suites, and relaxation settings.

For WA-SpecDec, the auxiliary next-frame latent prediction loss is weighted by $\lambda=0.1$. 
The world-aware signal is extracted from the frozen Wan 2.2 VAE encoder, which we use only as a latent compressor rather than invoking the Wan 2.2 video generation pipeline. 
Given the current RGB observation $I_t$, the frozen Wan 2.2 VAE encoder produces a spatial latent map $Z_t \in \mathbb{R}^{B \times H_z \times W_z \times C_{\mathrm{vae}}}$, which serves as the VAE feature space for the World-Aware Bias module. 
The latent map is tokenized, linearly projected, and processed by a four-layer spatial VideoEncoder before being spatially aligned to the VLA visual patch grid. 
The auxiliary VideoHead is used only during training and is discarded at inference, so inference does not run next-frame prediction, VAE decoding, or video generation.

For speculative tree decoding, we set the maximum number of tree nodes to 50, the tree depth to 4, and use the top-8 tokens to construct the draft tree. 
The same draft and verifier settings are used for Spec-VLA and WA-SpecDec under each relaxed acceptance rule. 
For the matched-speedup comparison in Table~\ref{tab:matched_speedup}, we set $\rho=10$ for the $\rho$-based verifier, while KERV and HeiSD use their default verifier settings. 
For the tolerance-sweep experiment, we evaluate $\rho \in \{5,10,15,20\}$.

Training uses a batch size of 32 and 1,000 warm-up steps. We use the OpenVLA-7B target policy. Wall-clock speedup is measured relative to autoregressive target decoding on the same hardware setup. The total computational budget for training, distillation, and evaluation is approximately 192 GPU hours, corresponding to 48 wall-clock hours on four NVIDIA RTX A6000 GPUs. The reported speedup includes all inference-time costs, including draft proposal, target verification, WAB computation, and key--value cache updates.
\section{Pseudocode}
\label{app:pseudocode}

Algorithm~\ref{alg:waspecdec} summarizes the training and inference procedure of WA-SpecDec. The input block collects the notation used throughout the algorithm, and the output block specifies the trained models and the continuous control sequence returned for a current query. At inference, the training-only auxiliary branch is removed, and WA-SpecDec uses only the current instruction--observation pair while applying the original relaxed acceptance rule without modification.

\begin{figure*}[t]
  \centering
  \includegraphics[width=\linewidth]{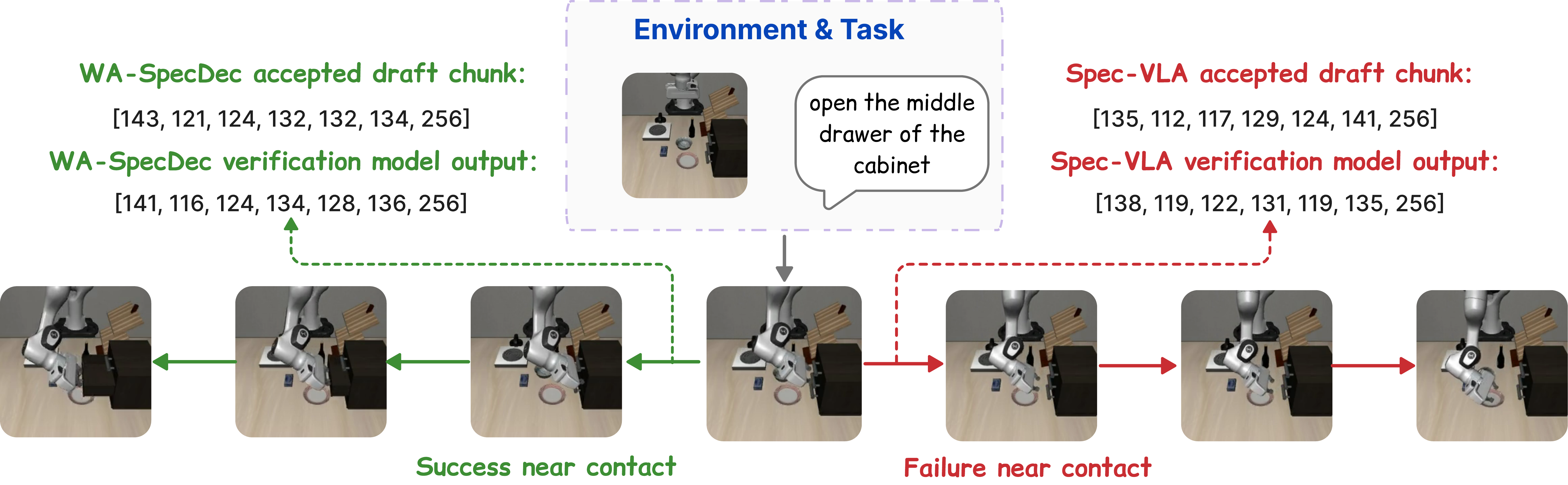}

\caption{
Case study of accepted action chunks under relaxed verification with \(\rho=10\) for the task ``open the middle drawer of the cabinet.'' Green denotes the WA-SpecDec accepted chunk and successful near-contact execution, while red denotes the Spec-VLA accepted chunk and near-contact failure.
}
  \label{fig:case_study}
\end{figure*}
\section{Case Study}
\label{app:casestudy}
Figure~\ref{fig:case_study} shows a representative near-contact case for the task ``open the middle drawer of the cabinet.'' Both Spec-VLA and WA-SpecDec produce an accepted action chunk under relaxed verification with $\rho=10$. Although both chunks satisfy the same token-distance acceptance criterion, their execution outcomes differ substantially. The Spec-VLA chunk is accepted in token space, but the resulting trajectory fails near the drawer-contact region. In contrast, WA-SpecDec completes the drawer-opening motion successfully.

This example illustrates a failure mode of token-distance-based relaxed acceptance in contact-sensitive states. A draft chunk that remains close to the verifier output in token space can still lead to task-ineffective motion when execution depends on fine-grained geometry near an object surface or handle. WA-SpecDec mitigates this issue by injecting world-aware information into the shared prefill hidden states, so that both draft proposal and target verification are conditioned on additional physical scene structure. In this case, the accepted WA-SpecDec chunk is better aligned with the near-contact execution requirement, improving reliability without changing the relaxed acceptance rule itself.

\end{document}